\pdfoutput=1

\documentclass[11pt]{article}

\usepackage[final]{acl}

\usepackage{times}
\usepackage{latexsym}
\usepackage[T1]{fontenc}

\usepackage[utf8]{inputenc}

\usepackage{microtype}
\usepackage{amsmath}
\usepackage{algorithmic}
\usepackage{algorithm}
\usepackage{inconsolata}

\usepackage{graphicx}
\usepackage{tcolorbox}
\usepackage{makecell}
\usepackage{longtable}
\usepackage{multicol}
\usepackage{multirow}
\usepackage{booktabs}
\usepackage{cite}
\usepackage{amsmath,amssymb,amsfonts}
\usepackage{algorithmic}
\usepackage{graphicx}
\usepackage{textcomp}
\usepackage{xcolor}
\usepackage{hyperref}
\title{Joint Enhancement of Relational Reasoning for Long-Context LLMs}

\author{
  Zhirui Chen$^\clubsuit$, Wei Shen$^\spadesuit$, Jiashui Huang$^\spadesuit$, Ling Shao$^\clubsuit$\thanks{Corresponding Author} \\
  $^\clubsuit$UCAS-Terminus AI Lab, University of Chinese Academy of Sciences, China \\
  $^\spadesuit$AI Lab, Terminus International, Terminus Group, China \\
  \texttt{\ chenzhirui23@mails.ucas.ac.cn,  ling.shao@ieee.org}
}

\begin{document}
\maketitle
\begin{abstract}
Despite significant progress, large language models (LLMs) still struggle with long contexts due to memory limitations and their inability to tackle complex and long-context tasks. Additionally, LLMs often suffer from a lack of transparency and are prone to producing hallucinations. To address these challenges, we propose \textbf{JERR}, a novel framework designed to enhance long-context comprehension via graph-based reasoning in LLMs. JERR integrates three key components: synopsis extraction, graph construction, and relational reasoning. First, synopsis is extracted by chunking text strategically, allowing the model to summarize and understand information more efficiently. Second, we build a directed acyclic graph (DAG) to resolve redundancy, ensuring logical consistency and clarity. Finally, we incorporate Monte Carlo Tree Search (MCTS) to help the model navigate complex reasoning paths, ensuring more accurate and interpretable outputs. This framework provides a novel solution that enables LLMs to handle extended contexts and complex reasoning tasks with improved reliability and transparency. Experimental results show that JERR consistently outperforms all baselines on the ROUGE and $F_1$ metrics, achieving the highest scores on the LLM-Rater evaluation.
\end{abstract}

\section{Introduction}

\label{sec:introduction}
In recent years, large language models (LLMs) have achieved significant achievements in natural language processing \citep{zhao2023survey}. However, transformer-based LLMs still face limitations in high-performance reasoning with long-context inputs due to constraints on memory usage and context window size \citep{liu2024lost, shi2023large}. Another challenge is that, although these models are pre-trained on extensive text corpora to respond coherently and appropriately to user inputs, they still struggle with complex knowledge-reasoning tasks \citep{hu2023survey,DBLP:conf/icml/0003AZPCW24,li2025explainable}.

We consider that the capability to handle long-context and graph-related tasks is critically important. Firstly, LLMs often struggle to provide accurate answers when key information is deeply embedded within lengthy contexts\citep{gandhi2024understanding} or when specialized knowledge extends beyond the pre-training corpora \citep{DBLP:conf/acl/JiangWL0L0Q24}, particularly with up-to-date information. Secondly, LLMs lack interpretability, transparency, and accountability, which increases the risk of producing hallucinations \citep{zhou2024cogmg}. Thirdly, few LLMs or frameworks operate in a human-like manner \citep{DBLP:conf/icml/LeeCFCF24,DBLP:conf/emnlp/LiHGBBLLQLOS024}. They do not think and respond in the thoughtful, deliberative way humans do when faced with challenging questions.

To address these three issues, researchers have proposed many methods to solve these problems recently, which can be roughly divided into two categories. The first approach focused on graph-related tasks by integrating LLMs with knowledge graphs \citep{DBLP:conf/iclr/LuoLHP24,DBLP:conf/icbk/JiangWLHCG24,DBLP:conf/iclr/SunXTW0GNSG24,DBLP:conf/emnlp/XuHC0STL0024}. These methods fine-tune baseline models using large datasets of generated triplets and reasoning paths \citep{liu2024few}. The second approach emphasizes training-free frameworks, which deliver high benchmark performance and broader generalizability, particularly for long-context processing, without compromising inference capabilities \citep{DBLP:conf/icml/LeeCFCF24,DBLP:conf/emnlp/LiHGBBLLQLOS024}. These frameworks also simplify knowledge updating \citep{DBLP:conf/acl/HanCBS24,wang2023knowledgpt}, allowing users to add new information to the knowledge base without retraining the model \citep{ibrahim2024survey}.

\begin{figure*}[ht]
\centering
\includegraphics[width=1\textwidth]{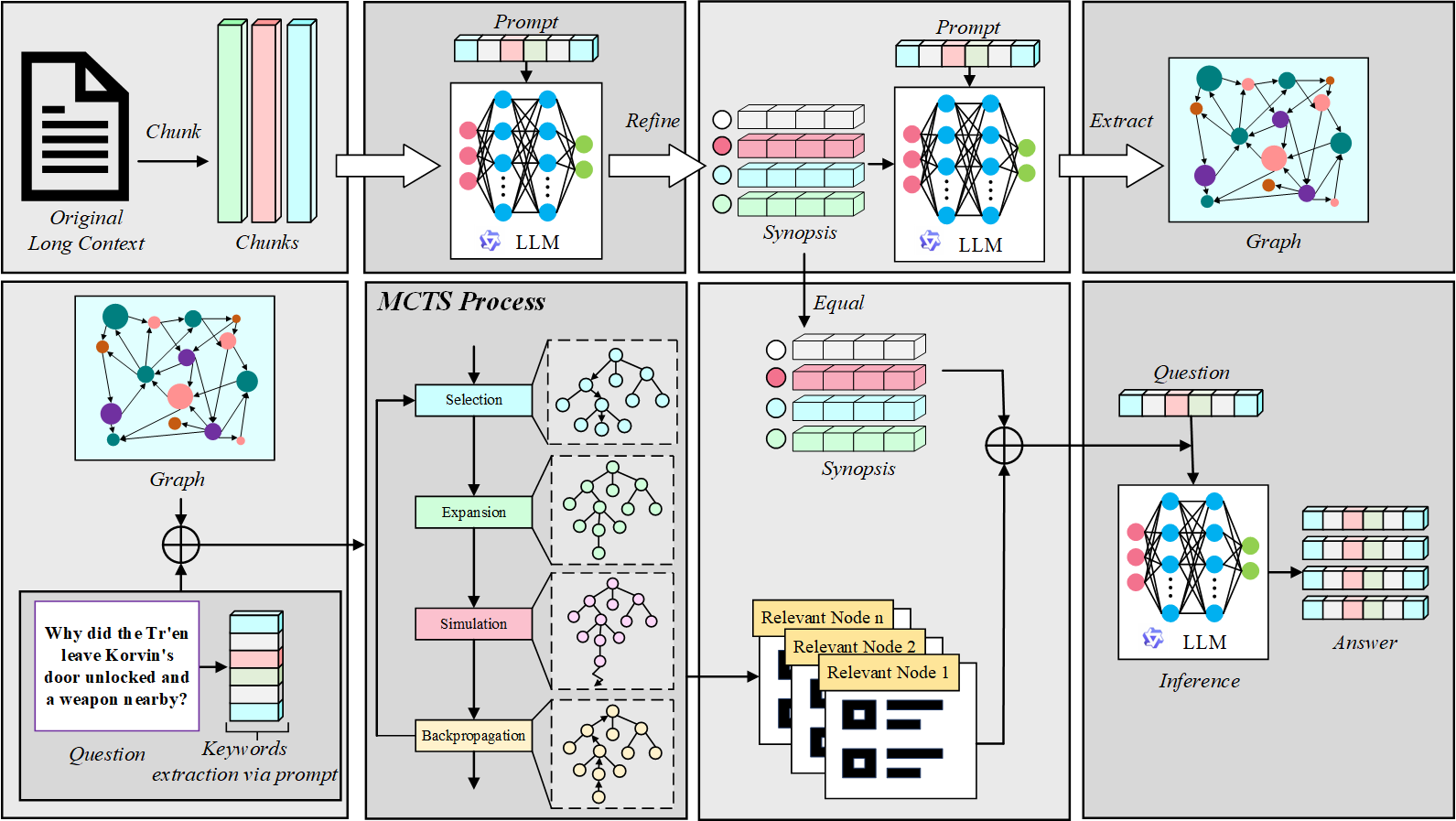} 
\caption{The overall framework of JERR consists of three main steps: 1) Given a question and a long context, LLMs are first prompted to segment the context into chunks. 2) Next, synopsis is generated through LLMs' prompts, followed by the construction of the graph. 3) Finally, reasoning is performed using relevant nodes, identified through the MCTS algorithm on the question and graph, along with the synopsis to generate the final answer.}
\label{fig1}
\end{figure*}

Given these challenges, we aim to design a generalized framework to enhance LLMs' capacity for processing long contexts and reasoning effectively. Our approach simulates the way humans utilize long-term memory by creating and memorizing summaries of key chunks as entities with interrelated attributes \citep{zhou2024cogmg}. This framework addresses long-context comprehension in three main stages: synopsis extraction, graph construction, and relational reasoning.

For synopsis extraction, we leverage LLMs' summarization prompts to strategically segment the context \citep{dong2023bridging,DBLP:conf/emnlp/HuangZ0VLP24}. In graph construction, we employ algorithms that address issues like entity redundancy \citep{DBLP:conf/nips/ChenZDH0H24} and attribute recurrence \citep{DBLP:conf/nips/NingL24}. This phase also involves building a directed acyclic graph (DAG) \citep{zhang2024diagram} to allow models to explore complex reasoning paths while maintaining logical coherence. Finally, in the relational reasoning phase, Monte Carlo Tree Search (MCTS) \citep{zhao2024large,gao2024interpretable,browne2012survey} enables precise retrieval of relevant nodes on the graph, enhancing the model's capacity to deliver accurate answers to complex queries.

To this end, we propose \textbf{JERR} (Figure \ref{fig1}), a \textbf{J}oint \textbf{E}nhancement of \textbf{R}elational \textbf{R}easoning for long-context LLMs.

\section{Related Work} \label{related work}

\subsection{Long-Context LLMs} Recent progress in long-context LLMs highlights training strategies that extend context windows to improve model performance \citep{beltagy2020longformer,DBLP:journals/access/Zhou23,DBLP:conf/nips/ZaheerGDAAOPRWY20,DBLP:conf/emnlp/AinslieLJOBZUGL23,DBLP:conf/iclr/ChenQTLL0J24,DBLP:conf/icml/DingZZXSX0Y24}. Optimized Transformer attention mechanisms have also been developed to efficiently manage long contexts without extensive fine-tuning \citep{DBLP:conf/iclr/PressSL22,DBLP:conf/icml/JinHYJLCC024,chen2023extending,DBLP:conf/iclr/XiaoTCHL24,han2024lm}. However, studies show that model performance often declines with longer inputs, even within the context limit, due to the influence of distracting elements \citep{liu2024lost,shi2023large}. Our work addresses these issues by improving the effective context length and filtering irrelevant content, avoiding the need for architectural changes or retraining.

\subsection{Retrieval} Retrieval-Augmented Generation (RAG) methods allow LLMs to access relevant information from large documents or text segments \citep{DBLP:conf/iclr/DinanRSFAW19,DBLP:conf/eacl/IzacardG21,park2023generative,lewis2020retrieval}. Research has examined retrieval granularity at levels such as tokens \citep{DBLP:conf/iclr/KhandelwalLJZL20}, entities \citep{DBLP:conf/emnlp/FevrySFCK20,DBLP:conf/iclr/JongZFSC22}, and chunks, with techniques ranging from traditional BM25 \citep{rasooli2015yara} to advanced learning-based approaches \citep{sachan2023questions}. Although RAG enhances retrieval accuracy, it struggles with complex queries due to limited decision-making. Our approach leverages relational reasoning in a graph-based framework to better retrieve and integrate relevant information for tasks requiring long-context understanding.{}

\subsection{Agent for Retrieval} Interactive agents have been employed to help LLMs navigate and process long texts. PEARL \citep{DBLP:conf/nips/ChenZDH0H24} uses iterative prompting to improve understanding, and Self-note \citep{lanchantin2024learning} integrates notes with documents for better reasoning. LongRAG \citep{jiang2024longrag} introduces a “long retriever” and a “long reader”, allowing the entire corpus to be processed into larger-sized units, which reduces the number of units needed during retrieval and alleviates the burden on the retriever. GraphRAG \citep{edge2024local} introduces a graph-based retrieval augmentation framework that enhances large language models by explicitly modeling entity relationships and semantic associations within knowledge sources. GraphReader \citep{DBLP:conf/emnlp/LiHGBBLLQLOS024} addresses these issues by structuring text into graphs and employing agents for autonomous exploration.

\section{Methodology}
This section details the design and implementation of our framework, JERR, an agent that integrates graph structures with long-context processing capabilities.

\subsection{Overview of the Approach}
As discussed in Section \ref{sec:introduction}, there remains a significant challenge in balancing long-context processing with strong reasoning capabilities in LLMs. To address this, we propose a three-step approach that combines synopsis extraction, graph construction, and graph-based reasoning.

For synopsis extraction, we use the \verb|autogen|\footnote{\href{https://github.com/microsoft/autogen}{https://github.com/microsoft/autogen}} package to segment the original long context into chunks. Each chunk is then refined into a synopsis through targeted prompts specifically designed for summarization.

For graph construction, we extract nodes and attributes via prompts of information atoms, core components and attributes extraction. Then we incorporate both exact and similar deduplication techniques. We utilize a Bloom Filter with Trie for exact match and SimHash for similarity-based deduplication. Subsequently, inspired by the core approach in \citeposs{shi2023large} work, we construct a Directed Acyclic Graph (DAG), which significantly enhances the efficiency of graph traversal, allowing the agent to conduct more focused exploration during the reasoning process. 

For graph-based reasoning, we identify the top-\emph{k} nodes most relevant to a given question via employing the Monte Carlo Tree Search (MCTS) algorithm. The resulting subset of relevant nodes, along with the extracted synopsis, is used to prompt the agent. This allows the agent to identify specific sections of the original text that should be revisited and replaced. Finally, the combination of the extracted synopsis and selected passages is used to generate a well-informed response to the query.

\subsection{Synopsis Extraction}
To extend the capacity of large language models (LLMs) for long-context processing, a foundational approach involves segmenting the context into manageable chunks through prompt-based or conventional chunking methods.

The \verb|autogen| chunking function allows for various customization parameters, including a maximum token limit per chunk, a specified chunking mode, an option to enforce breaks at empty lines, and an overlap line count to ensure continuity between consecutive chunks. Given an input context $T$, the function can be expressed as follows:

\begin{equation}
chunk\left( {T} \right) =\left\{ t_1, t_2, ..., t_{\mathrm{n}} \right\} 
\end{equation}
In the chunking phase, the input text is divided into a set of segments, denoted as $\left\{ t_1, t_2, ..., t_{\mathrm{n}} \right\}$. This segmentation is followed by a synopsis extraction prompt (in $\S$\ref{A.1}) with specific parameter settings, designed to facilitate efficient storage and rapid retrieval of the most relevant information from the extended context. This process provides a structured approach to distilling key information from each chunk. The resulting set of synopsis segments can be represented as:
\begin{equation}
S=\left\{ s_1,s_2,...,s_{\mathrm{n}} \right\} =\underset{i=1}{\overset{\mathrm{n}}{U}}p_{\mathrm{syn}}\left( t_{\mathrm{i}} \right)
\end{equation}
where $p_{\mathrm{syn}}$ denotes the synopsis extraction prompt. Each synopsis segment $s_{\mathrm{n}}$ thus captures the essential content of its corresponding chunk $t_i$.

\subsection{Graph constructing}
To maximize the effectiveness of synopsis, we developed a graph construction method that structures an information database, capturing key entities and relationships within the text. This approach provides a robust pipeline for entity deduplication and the construction of a Directed Acyclic Graph (DAG) that effectively encapsulates semantic relationships across the text corpus.

Initially, entities and core elements are extracted from synopsis using the corresponding prompts  (in $\S$\ref{A.2} and $\S$\ref{A.3}). Then we process exact deduplication, where each entity is passed through a Bloom Filter\footnote{\href{https://pypi.org/project/pybloom/}{https://pypi.org/project/pybloom/}} and Trie structure to retain only unique entities. The set of exact deduplicated elements is defined as:
\begin{equation}
C=\left\{ {c}_\mathrm{i}\left| c=ddp_{\mathrm{exact}}\left( s_{\mathrm{j}} \right) ,\,\,\forall s_{\mathrm{j}}\in S \right. \right\}
\end{equation}
where ${c}_\mathrm{i}$ represents the result of deduplication for each synopsis segment $s_{\mathrm{j}}$ and $ddp_{\mathrm{exact}}$ denotes the exact deduplication function.

A second deduplication step is performed on the unique entities from the previous stage using the SimHash\footnote{\href{https://pypi.org/project/simhash/}{https://pypi.org/project/simhash/}} algorithm. SimHash approximates similarity between entities by generating hash values, which are compared to determine near-duplicates. Entities with a bitwise difference below a set threshold (e.g., three bits) are considered similar and are combined. Similarly, the set of similar deduplicated elements is denoted as:
\begin{equation}
V=\left\{ {v}_\mathrm{i}\left| v=ddp_{\mathrm{sim}}\left( c_{\mathrm{j}}, \theta \right) ,\,\,\forall c_{\mathrm{j}}\in C \right. \right\} 
\end{equation}
where ${v}_\mathrm{i}$ denotes the final set of candidate nodes after two rounds of deduplication, $\theta$ is the similarity threshold and $ddp_{\mathrm{sim}}$ represents the SimHash-based deduplication function.

The final step involves constructing graph. In this stage, unique entities from deduplication stages are iteratively added as nodes in the graph. Each node is annotated with the corresponding synopsis derived from the segmented context. For each pair of entities within a chunk, a directed edge is proposed based on the existence of relational attributes between them. These relational attributes are determined using an edge attribute generation prompt (in $\S$\ref{A.4}), and the resulting set of attributes can be expressed as:
\begin{equation}
\scalebox{0.8}{ 
$E=\left\{ \left( v_{\mathrm{i}},\,\,v_{\mathrm{j}} \right) \left| \,\,prompt_{\mathrm{ea}}\left( v_{\mathrm{i}},\,\,v_{\mathrm{j}} \right) \ne 0,\,\,\forall v_{\mathrm{i}},v_{\mathrm{j}}\in V \right. \right\}$}
\end{equation}
Thus, graph $G$ is defined as $G=(V,E)$.

\subsection{Graph-based Reasoning}
After constructing the graph $G$, we identify the most relevant nodes using prompt-extracted keywords from query and the Monte Carlo Tree Search (MCTS) as a graph search mechanism. Upon retrieving the relevant nodes, a prompt-based mechanism guides the agent in selecting key chunks from the original context. The agent then uses both the retrieved nodes and the selected context chunks to perform an inference step and generate a response to the user's query.

The Monte Carlo Tree Search (MCTS) mechanism searches for relevant nodes by iteratively simulating paths from a root node set $V$ to potential child nodes within the synopsis graph. During the selection phase, it traverses down the tree, choosing child nodes with the highest win/visit ratio to maximize exploration of promising nodes. In the expansion phase, it adds unexplored neighbor nodes as children if the selected node lacks children. The simulation phase computes scores by evaluating keyword matches between the query and the nodes in the simulated path, approximating relevance based on keyword overlap. Finally, in backpropagation, these scores are updated along the path to the root, guiding the search toward nodes that yield high relevance scores in future iterations. Given a specified number of returned nodes $k$, graph $G$ and user query $q$, the resulting set of candidate nodes after the MCTS algorithm can be defined as:
\begin{equation}
R=MCTS\left( G,\,\,q,\,\,k \right)
\end{equation}
The above MCTS process is detailed as $\mathrm{\S}$\ref{B}.

\section{Experiment}
In this section, we describe the experimental setup and present our main results with an ablation study, case study of our proposed approach (cost analysis is in $\S$\ref{C.2}).

\subsection{Experiment Setup}

\paragraph{Dataset} We conduct experiments on three types of long-context QA benchmarks, QuALITY, MuSiQue and NarrativeQA, shown in Table~\ref{tab1}. See more details in $\S$\ref{C.1}.

\begin{table}[htbp]
\centering
\caption{The statistic of benchmark in our experiments to conduct three types of evaluation on our agent}
\scalebox{0.63}{ 
\begin{tabular}{ccccc}
\Xhline{1pt}
\textbf{Task}              & \textbf{Dataset} & \textbf{Avg Tokens} & \textbf{Max Tokens} & \textbf{Samples} \\ \hline
\textbf{Multi-choice   QA} & QuALITY          & 4.1k                 & 6.0k                 & 230              \\ 
\textbf{Multi-hop QA}      & MuSiQue          & 15.5k               & 16.0k               & 200              \\ 
\textbf{Single-hop QA}     & NarrativeQA      & 29.7k               & 63.7k               & 200              \\ \Xhline{1pt}
\end{tabular}%
}
\label{tab1}
\end{table}

\paragraph{Baselines} 
We evaluate the proposed approach against several established baselines:

\textbf{Retrieval Augmented Generation: } As discussed in Section \ref{related work}, retrieval-augmented generation (RAG) is a widely used approach for accessing extensive text collections beyond the LLM's context window. In our study, we implement two RAG variants: one utilizing \citet{bm25} and the other leveraging neural retrieval with the Qwen API's \citet{aliyunembedding} to retrieve the chunks most relevant to the user's query. The retrieved chunks are then processed by the \citet{qwenplus} model to generate answers. 

\textbf{Long-context LLM: } Given the cost constraints for handling extensive long-context benchmarks, we choose not to use GPT-4, instead selecting the high-performance qwen-plus-128k model to directly process the entire input passage. This choice aligns well with our selected datasets, which fit within the qwen-plus-128k input window capacity.

\textbf{Agent-based Method: } For agent-based retrieval, we use ReadAgent \citep{DBLP:conf/icml/LeeCFCF24}, a model inspired by human reading processes that performs interactive retrieval and reading, demonstrating effective handling of long-context QA tasks. We replace the PALM-2L model used in the ReadAgent study with the qwen-plus-128k model. GraphReader \citep{DBLP:conf/emnlp/LiHGBBLLQLOS024} was not incorporated as a comparative baseline in the present study due to the absence of public source code, thus lacking its operational integration with qwen-plus-128k. We've clarified the differences among JERR, GraphReader and GraphRAG in $\S$\ref{D}. While GraphRAG \citep{edge2024local} and LongRAG \citep{jiang2024longrag}, as excellent agent-based methods, can be compared during experiments.

\paragraph{Evaluation Settings} To assess the long-context summarization capabilities of our model, we employ several automatic evaluation metrics, including ROUGE (R-1, R-2, and R-L) and $F_1$ score. While these automatic metrics offer high efficiency, the accuracy may vary depending on the model's response format. To address this, we incorporate LLM Raters within ReadAgent to evaluate answer correctness by prompting a comparison to ground truth responses. The LLM Raters include LR-1 (strict version) and LR-2 (permissive version). The specific prompts of them can be seen in \citep{DBLP:conf/icml/LeeCFCF24}.
{}
\paragraph{Implementation Details} In the RAG baseline, chunk sizes are tailored to the specific characteristics of each dataset: 2000 tokens for NarrativeQA and MuSiQue, and 600 tokens for QuALITY, to account for differences in passage lengths. The top-$k$ parameter in MCTS is set to 5, a choice informed by ablation study results. To identify relevant pages across datasets, we adapt a look-up prompt approach inspired by ReadAgent \citep{DBLP:conf/icml/LeeCFCF24}, applying dataset-specific configurations to effectively address questions in QuALITY, NarrativeQA, and MuSiQue, as outlined in $\S$\ref{A.5} and $\S$\ref{A.6}.

\begin{table}
\centering
\caption{Comparison of Different Methods on \textbf{QuALITY} Datasets}
\scalebox{0.69}{
\begin{tabular}{cc}
\hline
\textbf{Method}           & \textbf{ACC}     \\ \hline
\textbf{BM25 Retrieval with qwen API}   &                  \\ \hline
Top-1                     & 58.68\%          \\
Top-2                     & 66.97\%          \\
Top-3                     & 73.21\%          \\
Top-4                     & 74.59\%          \\
Top-5                     & 78.00\%          \\
Top-6                     & 79.91\%          \\ \hline
\textbf{Neural Retrieval with qwen API} &                  \\ \hline
Top-1                     & 66.30\%          \\
Top-2                     & 73.92\%          \\
Top-3                     & 78.14\%          \\
Top-4                     & 79.43\%          \\
Top-5                     & 81.30\%          \\
Top-6                     & 83.32\%          \\ \hline
\textbf{qwen-plus-128k}   & 84.80\%          \\ \hline
\textbf{ReadAgent with qwen API}        & 83.80\%          \\ \hline
\textbf{LongRAG with qwen API}               & 84.91\% \\ \hline
\textbf{GraphRAG with qwen API}               & 85.02\% \\ \hline
\textbf{JERR with qwen API}               & \textbf{86.39\%} \\ \hline
             
\end{tabular}
}
\label{tab2}
\end{table}

\begin{table*}
\centering
\caption{Comparison of Different Methods on \textbf{MuSiQue} and \textbf{NarrativeQA} Datasets}
\scalebox{0.7}{
\begin{tabular}{c|cccccccccccc}
\hline
\textbf{Dataset}               & \multicolumn{6}{c|}{\textbf{MuSiQue}}                                                                                            & \multicolumn{6}{c}{\textbf{NarrativeQA}}                                                                    \\ \hline
\textbf{Method}                & \textbf{LR-1}  & \textbf{LR-2}  & \textbf{R-1}     & \textbf{R-2}     & \textbf{R-L}     & \multicolumn{1}{c|}{\textbf{\textit{F$_1$}}}      & \textbf{LR-1}  & \textbf{LR-2}  & \textbf{R-1}     & \textbf{R-2}     & \textbf{R-L}     & \textbf{\textit{F$_1$}}      \\ \hline
BM25 Retrieval with qwen API                & \multicolumn{12}{c}{}                                                                                                                                                                                                                          \\ \hline
Top-1                          & 0.175          & 0.275          & 0.147         & 0.094         & 0.139         & \multicolumn{1}{c|}{0.253}         & 0.165          & 0.300          & 0.100         & 0.076         & 0.104         & 0.182           \\
Top-2                          & 0.260           & 0.380           & 0.184          & 0.149         & 0.181         & \multicolumn{1}{c|}{0.291}         & 0.255           & 0.430           & 0.134          & 0.108         & 0.122         & 0.194         \\
Top-3                          & 0.325          & 0.47           & 0.161         & 0.116          & 0.153          & \multicolumn{1}{c|}{0.251}         & 0.345          & 0.520           & 0.138         & 0.112          & 0.135          & 0.221         \\
Top-4                          & 0.355          & 0.505          & 0.172         & 0.137         & 0.168         & \multicolumn{1}{c|}{0.233}         & 0.340          & 0.575          & 0.143         & 0.103         & 0.131         & 0.217          \\
Top-5                          & 0.425          & 0.535          & 0.159         & 0.112         & 0.156         & \multicolumn{1}{c|}{0.267}         & 0.420          & 0.585          & 0.150         & 0.143         & 0.150         & 0.238         \\
Top-6                          & 0.41           & 0.525          & 0.160         & 0.109         & 0.154         & \multicolumn{1}{c|}{0.286}         & 0.445           & 0.645          & 0.155         & 0.129         & 0.150         & 0.237         \\ \hline
Neural Retrieval with qwen API & \multicolumn{12}{c}{}                                                                                                                                                                                                                          \\ \hline
Top-1                          & 0.185          & 0.295          & 0.072         & 0.068         & 0.070         & \multicolumn{1}{c|}{0.235}         & 0.290          & 0.410          & 0.075         & 0.053         & 0.069         & 0.156          \\
Top-2                          & 0.275          & 0.405          & 0.079         & 0.062         & 0.05         & \multicolumn{1}{c|}{0.269}         & 0.340          & 0.525          & 0.077         & 0.056         & 0.070         & 0.160         \\
Top-3                          & 0.335          & 0.450           & 0.088         & 0.068         & 0.084         & \multicolumn{1}{c|}{0.231}           & 0.390          & 0.570           & 0.081         & 0.051         & 0.073         & 0.162         \\
Top-4                          & 0.395          & 0.535          & 0.087          & 0.074         & 0.083         & \multicolumn{1}{c|}{0.288}         & 0.445          & 0.635          & 0.087          & 0.059         & 0.079         & 0.178         \\
Top-5                          & 0.425          & 0.565          & 0.094         & 0.076         & 0.091         & \multicolumn{1}{c|}{0.295}         & 0.445          & 0.645          & 0.089         & 0.055         & 0.081         & 0.166         \\
Top-6                          & 0.405          & 0.525          & 0.091         & 0.074         & 0.089         & \multicolumn{1}{c|}{0.272}         & 0.475          & 0.685          & 0.091         & 0.061         & 0.083         & 0.165         \\ \hline
qwen-plus-128k                 & 0.425          & 0.515          & 0.095         & 0.075         & 0.091         & \multicolumn{1}{c|}{0.328}         & 0.515          & 0.735          & 0.104         & 0.066         & 0.094         & 0.251           \\ \hline
ReadAgent with qwen API                      & 0.405           & 0.530          & 0.206         & 0.169         & 0.201         & \multicolumn{1}{c|}{0.448}         & 0.520           & 0.735          & 0.230         & 0.209         & 0.212         & 0.247         \\ \hline
LongRAG with qwen API                      & 0.415           & 0.545          & 0.209         & 0.174         & 0.203         & \multicolumn{1}{c|}{0.473}         & 0.520           & 0.735          & 0.228         & 0.204         & 0.201         & 0.249         \\ \hline
GraphRAG with qwen API                      & 0.410           & 0.550          & 0.211         & 0.185         & 0.205         & \multicolumn{1}{c|}{0.488}         & 0.525           & 0.745          & 0.221         & 0.205         & 0.207         & 0.254         \\ \hline
JERR with qwen API                         & \textbf{0.455} & \textbf{0.595} & \textbf{0.226} & \textbf{0.212} & \textbf{0.218} & \multicolumn{1}{c|}{\textbf{0.505}} & \textbf{0.540} & \textbf{0.760} & \textbf{0.234} & \textbf{0.215} & \textbf{0.216} & \textbf{0.269} \\ \hline
\end{tabular}
}
\label{tab3}
\end{table*}

\subsection{Overall Performance Comparison}
The comparative results of the four baseline methods and the proposed JERR framework on multiple-choice question answering, multi-hop, and single-hop long-context question-answering benchmarks are presented in Table \ref{tab2} and Table \ref{tab3}.

\paragraph{RAG Methods}
BM25 and Neural Retrieval approaches demonstrate varying effectiveness across different benchmark tasks. In the QuALITY benchmark (Table \ref{tab2}), both retrieval methods underperform compared to direct reading approaches, suggesting that chunking and ranking processes may introduce unnecessary complexity. The retrieval methods show consistent accuracy improvements up to Top-5/Top-6 chunks, after which returns diminish.

BM25 retrieval consistently outperforms neural retrieval in lexical-based metrics (ROUGE-1, ROUGE-2, ROUGE-L) across MuSiQue and NarrativeQA benchmarks in Table \ref{tab3}. For instance, in MuSiQue, BM25 achieves an R-1 score of 0.159 at Top-5, significantly higher than neural retrieval's 0.094. This performance gap highlights BM25's strength in capturing lexical overlap.

For LR metrics, both retrieval methods exhibit similar patterns: starting with low Top-1 scores, they improve gradually with additional chunks, typically peaking at Top-5 or Top-6. In MuSiQue, both methods reach comparable maximum scores (LR-1: 0.425, LR-2: ~0.55), while in NarrativeQA, neural retrieval achieves peak performance with Top-6 chunks (LR-1: 0.475, LR-2: 0.685). However, increasing retrieval chunks beyond these points shows diminishing returns, indicating a trade-off between comprehensive coverage and precision.

\paragraph{Long-Context LLMs}
qwen-plus-128k demonstrates distinct capabilities across different benchmark tasks. In the QuALITY benchmark, the model achieves strong accuracy through direct question-answering without context segmentation, effectively avoiding the "lost in the middle" issue \citep{liu2024lost} when input lengths fall within its context window.

For complex summarization tasks in MuSiQue and NarrativeQA, qwen-plus-128k shows mixed performance patterns. While achieving relatively lower ROUGE scores (R-1: 0.095-0.104, R-2: 0.066-0.075, R-L: 0.091-0.094) compared to retrieval-based approaches, the model demonstrates stronger performance in LR metrics. Specifically, it achieves LR-1 scores of 0.425-0.515 and LR-2 scores of 0.515-0.735 across these benchmarks, indicating robust semantic understanding of long-form content.

The \textit{F$_1$} scores (0.328 in MuSiQue, 0.251 in NarrativeQA) suggest balanced precision-recall performance, though generally lower than BM25 retrieval approaches. However, a key advantage of qwen-plus-128k lies in its ability to process entire documents holistically, achieving comparable recall to chunking-based approaches without the overhead of document segmentation and selection.

\paragraph{Agent-based Methods}

Our framework, JERR, consistently demonstrates superior performance across diverse benchmarks by leveraging agent-based approaches for reasoning and information retrieval. In the QuALITY benchmark (Table \ref{tab2}), JERR outperforms other baselines in multiple-choice question-answering tasks. By employing the Monte Carlo Tree Search (MCTS) algorithm for graph construction and exploration, JERR achieves robust graph comprehension and precise relational reasoning. This enables it to efficiently retrieve and synthesize relevant information, minimizing information loss. In contrast, methods such as ReadAgent, which rely on compressing context into gist memories, face limitations in accuracy due to significant information loss during compression.

For the MuSiQue benchmark (Table \ref{tab3}), JERR demonstrates a clear advantage in long-range recall tasks, achieving superior LR-1 and LR-2 scores of 0.455 and 0.595, respectively, compared to ReadAgent's 0.405 and 0.53. Additionally, JERR surpasses ReadAgent across automatic evaluation metrics, including ROUGE and \textit{F$_1$} scores, showcasing its effectiveness in handling multi-sentence reasoning tasks.

In the NarrativeQA benchmark (Table \ref{tab3}), JERR achieves the highest ROUGE, \textit{F$_1$}, and LR scores, excelling in both detailed lexical matching and maintaining narrative coherence. This highlights its ability to adapt to complex narrative structures, effectively preserving key elements from the source text while ensuring a cohesive understanding of the storyline. Across all datasets, JERR is a reliable and versatile approach for long-text reasoning and retrieval tasks.

The results suggest that JERR's graph-based approach offers an advantage in processing and reasoning over extended narrative information, making it a highly effective solution for retrieval-based tasks on narrative datasets.

\subsection{Ablation Study}
\paragraph{The Effect of MCTS}
 ~

A critical step in JERR's processes is the MCTS algorithm setting. Thus, in Table \ref{tab5}, we compare JERR's performance using the MCTS algorithm with that of the PageRank algorithm on the QuALITY benchmark. The results clearly show that JERR achieves better performance with MCTS, boosting accuracy by approximately 5\% and demonstrating a notable improvement.

\begin{table}[htbp]
\caption{w. / w.o. MCTS accuracy Comparison on QuALITY of JERR}
\begin{center}
\scalebox{0.7}{
\begin{tabular}{l|l}
\hline
Method                         & Accuracy \\ \hline
JERR (via MCTS algorithm)     & 86.39\%  \\ \hline
JERR (via PageRank algorithm) & 81.69\%  \\ \hline
\end{tabular}
}
\label{tab5}
\end{center}
\end{table}
\paragraph{Impact of the Number of Relevant Nodes}
 ~
 
\begin{figure}[htbp]
    \centering
    \includegraphics[width=0.45\textwidth]{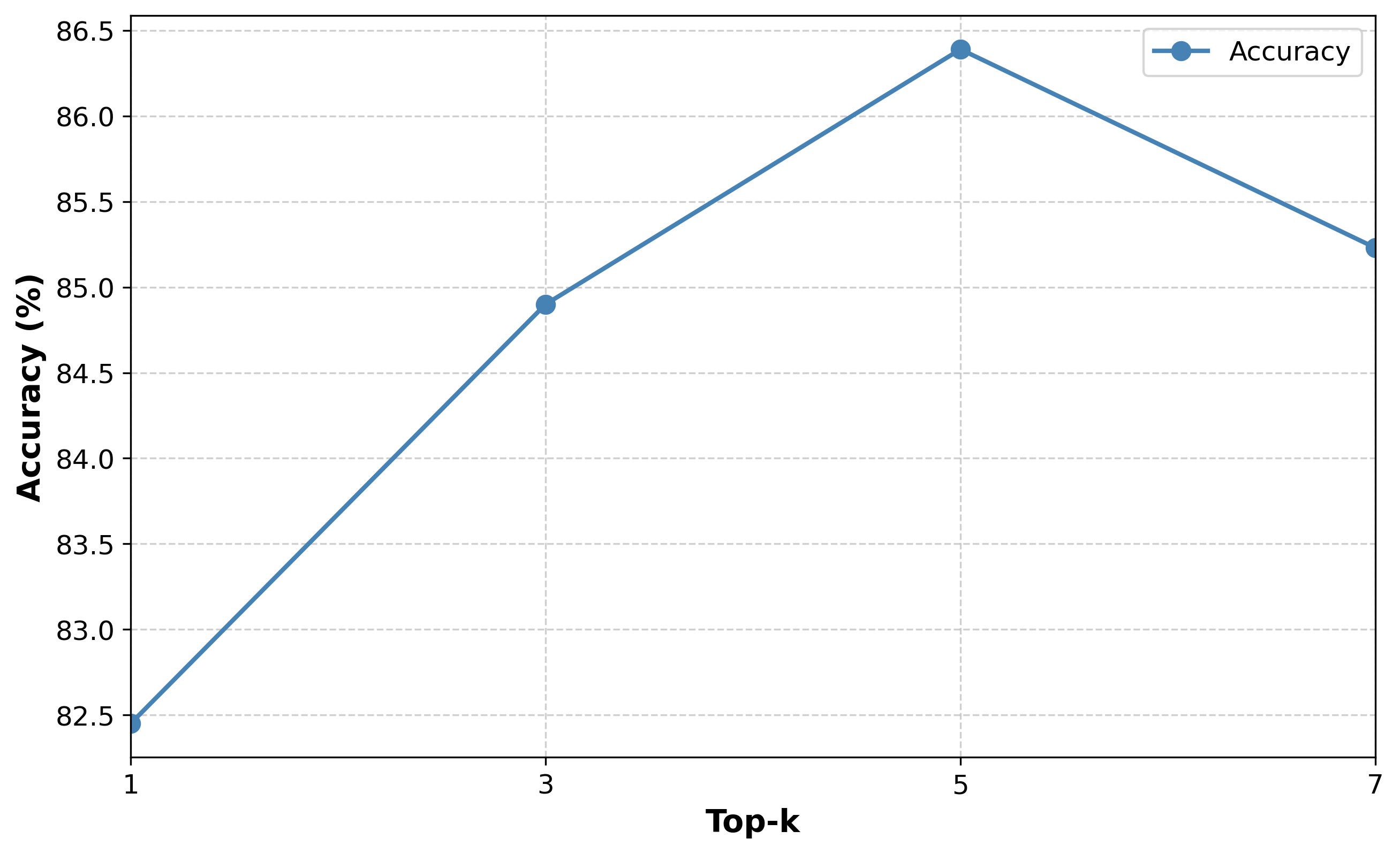}  
    \caption{Performance of JERR with different top-k relevant nodes text on QuALITY.}
    \label{fig2}
\end{figure}

Another essential factor in JERR's performance is the selection of the top-k values. We conducted experiments with various top-k values on the QuALITY dataset to assess their impact. As shown in Figure \ref{fig2}, performance improves as top-k values increase, peaking at an optimal value of 5, which we adopt as the default setting. Beyond this point, however, performance declines, likely due to an overload of retrieved information, which may hinder the model's inference capability.

\begin{figure}[htbp]
  \centering
  \includegraphics[width=0.5\textwidth]{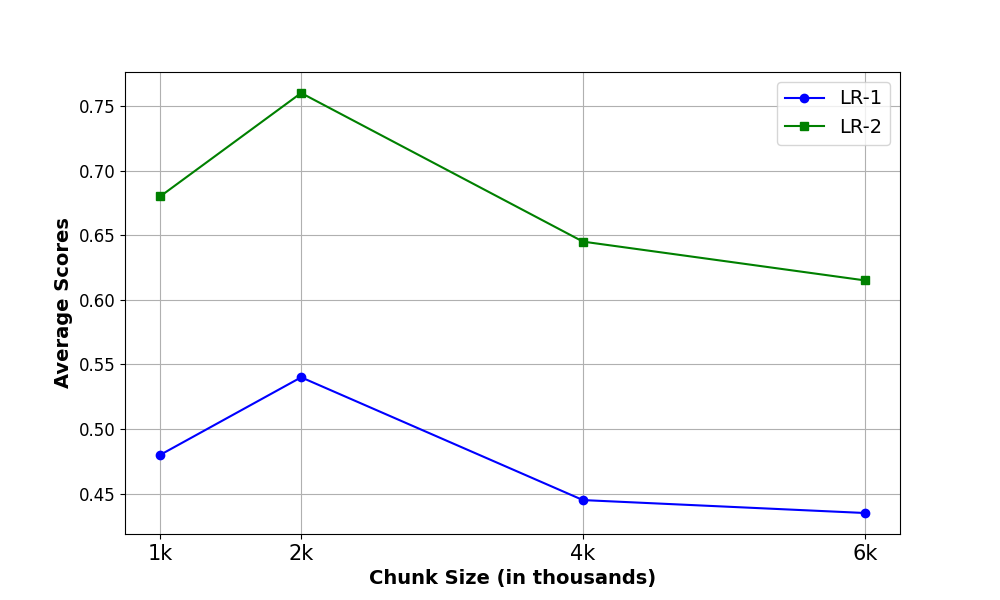}  
  \caption{Performance of JERR with different chunk size on NarrativeQA.}
  \label{fig3}
\end{figure}

\begin{figure*}[!ht]
\centering
\includegraphics[width=0.91\textwidth]{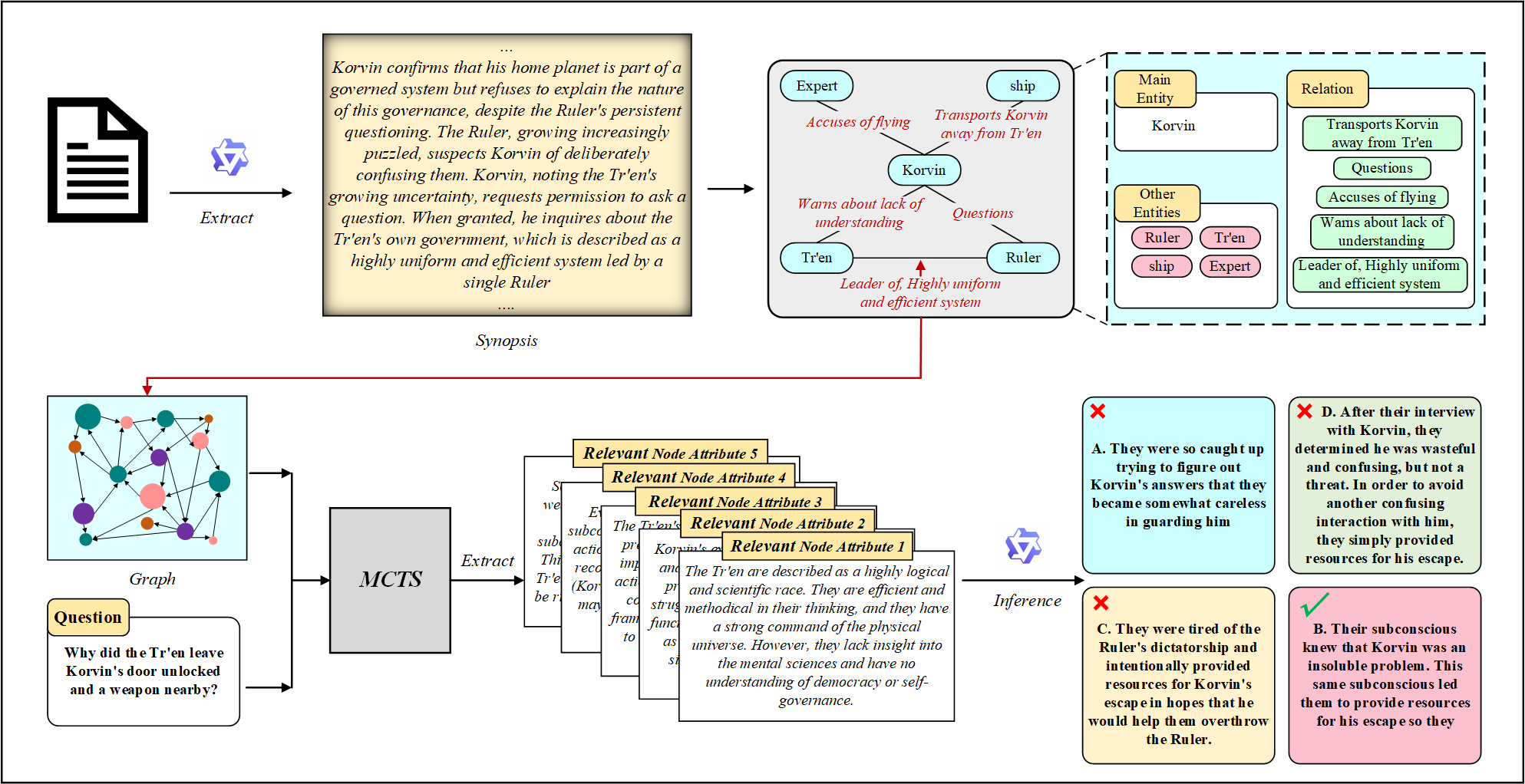} 
\caption{Study of case from the second passage of QuALITY dev set with the corresponding question (1 out of 9)}
\label{fig5}
\end{figure*}

\begin{figure}[htbp]
    \centering
    \includegraphics[width=0.45\textwidth]{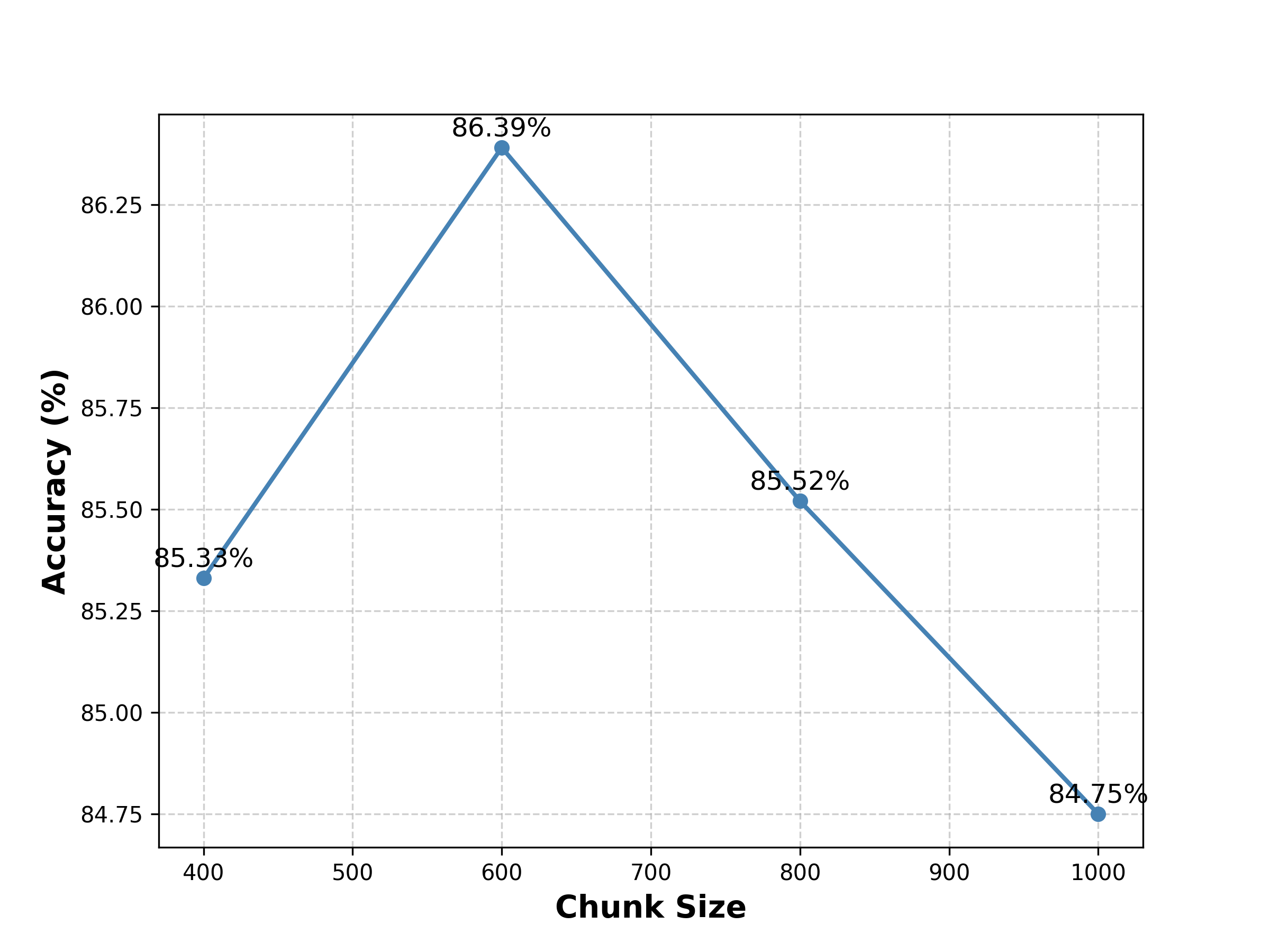}  
    \caption{Performance of JERR with different chunk size on QuALITY.}
    \label{fig4}
\end{figure}

\paragraph{Impact of Chunk Size}

For different setting of chunk size, we conduct experiments comparing LR-1 and LR-2 to evaluate JERR's performance on NarrativeQA dataset. As shown in Figure \ref{fig3}, the best performance is achieved with chunk size of 2k. It is evident that performance declines when the chunk size is set too large, which we attribute to model's inability to capture all details within longer chunks. Smaller chunk sizes allow for more accurate extraction of the synopsis. Therefore, we select chunk size of 2k as default.

In the QuALITY dataset, due to its shorter average context length, we use chunk sizes under 1000 tokens to maintain an optimal number of segments. Consistent with our findings on the NarrativeQA dataset, Figure \ref{fig4} demonstrates that accuracy decreases when chunk size exceeds an optimal threshold. Peak performance is observed at a chunk size of 600, which we therefore set as the default for QuALITY.

\subsection{Case Study}
This section demonstrates JERR's workflow through a case study from QuALITY, analyzing a 4,168-word passage about Korvin's interactions with the Tr'en society, which lacks understanding of democratic principles and mental sciences.

As illustrated in Figure \ref{fig5}, the original passage undergoes refinement into synopsis chunks that preserve essential information. JERR then constructs a knowledge graph where nodes represent key concepts and edges indicate their relationships, enabling systematic mapping of textual information for efficient analysis.

JERR processes the question: \textit{"Why did the Tr'en leave Korvin's door unlocked and a weapon nearby?"} using Monte Carlo Tree Search (MCTS), which simulates various paths within the graph to identify the most pertinent nodes for answering the question.

Based on the MCTS traversal, relevant nodes (highlighted in red boxes) are extracted, encompassing critical concepts including the Tr'en's democratic incomprehension, their inability to consciously resolve problems outside their governance model, and subconscious actions stemming from their mental limitations.

The framework synthesizes these nodes to generate an inference, suggesting that the Tr'en's subconscious influenced their decision to leave Korvin's door unlocked, driven by their desire to eliminate the \textit{"problem"} he represented due to cognitive dissonance within their society.

The generated inference aligns with the correct answer (marked with a green check mark), confirming that the Tr'en subconsciously recognized Korvin as an insoluble problem and facilitated his escape to avoid further societal contradiction.

\subsection{Comparison to GraphReader on GPT-4}

We've conducted Experiment, which has the same settings in the paper of GraphReader, on MusiQUE and NarrativeQA to compare JERR with GraphReader and other baselines (e.g., ReadAgent, Pearl, LongRAG and GraphRAG) using GPT-4-128k. The table is presented in additional TABLE \ref{tab7} as below (the statistics in the table is aligned with Table 2 in GraphReader paper, MQ and NR stand for MuSiQue and NarrativeQA, respectively).

The result demonstrates that JERR outperforms GraphReader and other baselines using the base model GPT-4, which shows great robustness and proves that JERR Framework doesn't depend on a specific LLM.

\begin{table*}
\centering
\caption{Comparison to GraphReader}
\begin{center}

\scalebox{0.85}{
\begin{tabular}{lcccccc}
\toprule
\textbf{Methods}         & \textbf{LR-1 (MQ)} & \textbf{LR-2 (MQ)} & \textbf{$F_1$ (MQ)}  & \textbf{LR-1 (NR)} & \textbf{LR-2 (NR)} & \textbf{$F_1$ (NR)}  \\
\midrule
BM25 (Top-1)    & 33        & 36.5      & 23.9     & 29.5      & 34.5      & 11.3     \\
BM25 (Top-3)    & 43.5      & 49.5      & 31.1     & 44.5      & 52.5      & 20.5     \\
Ada-002 (Top-1) & 34.5      & 37        & 26.6     & 37.5      & 46.5      & 15.5     \\
Ada-002 (Top-3) & 40        & 45.5      & 32.1     & 45.5      & 53        & 19.5     \\
GPT-4-128k      & 52        & 59.5      & 42.7     & 63.5      & 77        & 29.4     \\
ReadAgent       & 54.5      & 61        & 45.1     & 63.5      & 75.5      & 18.9     \\
Pearl           & 45        & 51.5      & 33.3     & 43.5      & 48        & 16.2     \\
LongRAG         & 49        & 54.5      & 40.3     & 60.5      & 69.2      & 27       \\
GraphRAG        & 46.5      & 56        & 31.2     & 52        & 66.5      & 23.1     \\
GraphReader     & 59        & 63.5      & 47.4     & 65        & 80        & 29.8     \\
\textbf{JERR}        & \textbf{60.5}  & \textbf{64}    & \textbf{49.2} & \textbf{68}    & \textbf{83}    & \textbf{30.1} \\
\bottomrule
\end{tabular}

}
\end{center}
\label{tab7}
\end{table*}

\section{Conclusion}
We have introduced JERR, a graph with MCTS algorithm based agent, designed to combine effective long context and relational reasoning graph smoothly in LLMs. JERR transfers the input long context into synopsis, builds DAGs and employs MTCS algorithm to search the relational nodes information, thus guaranteeing high performance on long-context tasks versus all the baselines.

\section{Limitations}
Our current framework JERR has so far been validated exclusively on QuALITY, MuSiQue and NarrativeQA. Experimental results demonstrate that JERR has high performance on such long-context benchmarks. However, JERR has limitations mainly on three aspects: 

For task scalability, although JERR performs well on the selected datasets, its ability to generalize to other reasoning tasks in different knowledge domains remains uncertain. Further validation on a broader range of datasets is necessary to assess its scalability.

For knowledge graph construction, the framework relies on the Qwen API to extract graph nodes, which helps automate knowledge representation. However, constructing a refined knowledge graph still requires a complex pipeline which can make large-scale and high-precision graph building a challenging task.

For task complexity, JERR applies a graph-based approach and Monte Carlo Tree Search (MCTS) to improve knowledge reasoning and retrieve relevant text more effectively. However, its effectiveness on simpler tasks has not been systematically evaluated. Future studies should compare its performance on both simple and complex tasks to provide a more complete assessment of its capabilities.



\bibliography{custom}

\newpage
~
\newpage
\appendix

\section{Prompts}
\subsection{Synopsis Extraction Prompt}
\label{A.1}
\tcbset{
    examplebox/.style={
        colframe=black,         
        colback=gray!10,        
        coltitle=white,         
        fonttitle=\bfseries,    
        boxrule=0.75mm,         
        arc=2mm,                
        width=\columnwidth,     
        left=5mm,               
        right=5mm,              
        top=3mm,                
        bottom=3mm              
    }
}

\begin{tcolorbox}[examplebox]
    Please transfer the following chunk of a passage into synopsis. \\
    Just give me a synopsis version. No extra explanation. \\
    Passage: \{SYNOPSIS TEXT\}
\end{tcolorbox}

\subsection{Entities Extraction Prompt}
\label{A.2}
\tcbset{
    examplebox/.style={
        colframe=black,         
        colback=gray!10,        
        coltitle=white,         
        fonttitle=\bfseries,    
        boxrule=0.75mm,         
        arc=2mm,                
        width=\columnwidth,     
        left=5mm,               
        right=5mm,              
        top=3mm,                
        bottom=3mm              
    }
}

\begin{tcolorbox}[examplebox]
    You are a summarizing agent. Given a chunk of paragraph, summarize it into information atoms.
     (Information Atoms: Brief statements represent the most fundamental, indivisible facts, covering ideas like propositions, theories, entities, concepts, and underlying aspects such as reasoning, cause-effect relationships, event sequences, social interactions, timelines, and similar elements.)
\end{tcolorbox}

\subsection{Elements Extraction Prompt}
\label{A.3}
\tcbset{
    examplebox/.style={
        colframe=black,         
        colback=gray!10,        
        coltitle=white,         
        fonttitle=\bfseries,    
        boxrule=0.75mm,         
        arc=2mm,                
        width=\columnwidth,     
        left=5mm,               
        right=5mm,              
        top=3mm,                
        bottom=3mm              
    }
}

\begin{tcolorbox}[examplebox]
    You are a keyword extracting agent. Given a chunk of paragraph of a story, extract only 3 core components. 
\\
    (Core components: The fundamental nouns (e.g., people, moments, occurrences, settings, quantities), verbs (e.g., activities), and adjectives (e.g., conditions, emotions) that are central to the story's progression.)
\end{tcolorbox}

\subsection{Edge Attributes Prompt}
\label{A.4}
\tcbset{
    examplebox/.style={
        colframe=black,         
        colback=gray!10,        
        coltitle=white,         
        fonttitle=\bfseries,    
        boxrule=0.75mm,         
        arc=2mm,                
        width=\columnwidth,     
        left=5mm,               
        right=5mm,              
        top=3mm,                
        bottom=3mm              
    }
}

\begin{tcolorbox}[examplebox]
    Based on the atomic facts: '\{INFORMATION ATOMS1\}' and '\{INFORMATION ATOMS2\}', what are the attributes between \{ELEMENT1\} and \{ELEMENT2\}? \\
    (Information Atoms: smallest, indivisible truths extracted from text chunks.)\\
    Use no more than three words to answer.

\end{tcolorbox}

\subsection{QuALITY Answer Prompt}
\label{A.5}
\tcbset{
    examplebox/.style={
        colframe=black,         
        colback=gray!10,        
        coltitle=white,         
        fonttitle=\bfseries,    
        boxrule=0.75mm,         
        arc=2mm,                
        width=\columnwidth,     
        left=5mm,               
        right=5mm,              
        top=3mm,                
        bottom=3mm              
    }
}

\begin{tcolorbox}[examplebox]
Read the following article and answer a multiple choice question.
For example, if (C) is correct, answer with "Answer: (C) ...". No extra explanation please.

Article:
\{CONTEXT\}

Question:
\{QUESTION\}
\{OPTIONS\}
\end{tcolorbox}

\subsection{NarrativeQA / MuSiQue Answer Prompt}
\label{A.6}
\tcbset{
    examplebox/.style={
        colframe=black,         
        colback=gray!10,        
        coltitle=white,         
        fonttitle=\bfseries,    
        boxrule=0.75mm,         
        arc=2mm,                
        width=\columnwidth,     
        left=5mm,               
        right=5mm,              
        top=3mm,                
        bottom=3mm              
    }
}

\begin{tcolorbox}[examplebox]

\{RELEVANT NODES TEXT \& SYNOPSIS\}

Question:
\{QUESTION\}

Answer the question based on the above relevant content and extracted synopsis. Your answer should be short and concise.

\end{tcolorbox}

\section{MCTS Process}
\label{B}
\paragraph{Initialize the Search Tree} We define the search tree nodes $\mathcal{T}$, where each tree node corresponds to an informational node within the graph. Each node's state $s_{\mathrm{j}}\in S$ is associated with a graph node $v_{\mathrm{j}}\in V$ in graph $G$. The visit count $N(s)$ represents the number of times a node's state $s_{\mathrm{j}}$ has been visited, while the total reward $W(s)$ reflects the cumulative reward accumulated by visits to node's state $s_{\mathrm{j}}$. The average reward of node's state $s_{\mathrm{j}}$, utilized during the selection stage, is calculated as:
\begin{equation}
{Q}\left( {s} \right) =\frac{{W}\left( {s} \right)}{{N}\left( {s} \right)}
\end{equation}

\paragraph{Selection} The selection strategy employs the Upper Confidence Bound (UCB) to balance exploration and exploitation. An exploration coefficient $\kappa$ is introduced to regulate this balance. The UCB for a node $s$ is defined as:
\begin{equation}
UCB\left( s \right) =Q\left( s \right) +\kappa \cdot \sqrt{\frac{lnN\left( parent\left( s \right) \right)}{N\left( s \right)}}
\end{equation}
The selection criterion then chooses the node with the highest $UCB\left( s \right)$ in the current set of tree nodes $\mathcal{T}$:

\begin{equation}
s_{next}=\mathop {argmax} \limits_{s\in \mathcal{T}}UCB\left( s \right) 
\end{equation}

\begin{algorithm}[htp]
    \caption{MCTS Relevant Nodes Extraction Algorithm}
    \label{alg1}
    \label{alg:MCTSKeywordExtraction}
    \renewcommand{\algorithmicrequire}{\textbf{Input:}}
    \renewcommand{\algorithmicensure}{\textbf{Output:}}

    \begin{algorithmic}[1]
        \REQUIRE Graph $G$, Query text $Q$, Start nodes $S$, Number of simulations $N$
        \ENSURE Top nodes based on keyword relevance

        \STATE Initialize root node $R$ with state $S[0]$
        \STATE Extract keywords $K$ from $Q$ as set of lowercase words

        \FOR{$i = 1$ to $N$}
            \STATE \textbf{Selection:} Set current node $n \leftarrow R$
            \WHILE{$n$ has children}
                \STATE Select the best child of $n$ based on win/visit ratio
                \IF{$n$ has no selectable children}
                    \STATE \textbf{break}
                \ENDIF
            \ENDWHILE

            \STATE \textbf{Expansion:} 
            \IF{$n$ has no children}
                \STATE Expand children for $n$ using neighbors in $G$
            \ENDIF

            \STATE \textbf{Simulation:} 
            \IF{$n$ has children}
                \STATE Set current node $n$ to best child of $n$
            \ENDIF
            \STATE Initialize score $\text{score} \leftarrow 0$
            \FOR{depth $d = 1$ to $10$}
                \IF{$n$ exists in $G$}
                    \STATE Increment score by matching keywords $K$ with state of $n$
                    \STATE Set $n$ to best child of $n$
                    \IF{$n$ has no children}
                        \STATE \textbf{break}
                    \ENDIF
                \ENDIF
            \ENDFOR

            \STATE \textbf{Backpropagation:}
            \WHILE{$n$ is not null}
                \STATE Update visits and wins for $n$ based on score
                \STATE Set $n$ to parent of $n$
            \ENDWHILE
        \ENDFOR

        \STATE Sort root children by win/visit ratio and return top nodes

        \RETURN Top relevant nodes based on MCTS exploration
    \end{algorithmic}
\end{algorithm}

\paragraph{Expansion} If the selected node $s_{next}$ has not reached a terminal state or attained a sufficient visit count, the expansion step continues by adding its corresponding graph node $v_{next} \in V$ to its neighboring nodes. The appropriate set of nodes for expansion is defined as:
\begin{equation}
\mathcal{E} \left( s_{next} \right) =\left\{ s^{\prime}\in \mathcal{N} \left( v_{next} \right) \cap s^{\prime}\notin \mathcal{T} \right\} 
\end{equation}

\paragraph{Simulation} In this stage, a random simulation is conducted from one of the newly expanded nodes $s^{\prime}$ to estimate the potential reward along this path. The reward for state $s^{\prime}$ is evaluated based on its relevance to the query, which is determined by the maximum simulation depth $d$ and a set of keywords extracted from the query. The reward function is given as:

\begin{equation}
r\left( s^{\prime} \right) =\underset{i=1}{\overset{d}{\Sigma}}\left| K\cap \mathcal{E} \left( s^{\prime} \right) \right| 
\end{equation}

\paragraph{Backpropagation} The final step of MCTS is the Backpropagation phase, where the reward $r\left( s^{\prime} \right)$ obtained from the simulation is propagated back through each node along the traversal path, updating both the total reward $W(s)$ and visit count $N(s)$ for each node:
\begin{equation}
N(s_{n})=N(s_{n-1})+1
\end{equation}
\begin{equation}
W(s_{n})=W(s_{n-1})+r\left( s^{\prime} \right)
\end{equation}
As a result, the set of $top-k$, ranked by relevance, is used as the set of retrieved nodes $R$:
\begin{equation}
R=\left\{ \left. \left( r_i,\,\,\frac{W_i}{N_i} \right) \right|i=1,2,...,k \right\} 
\end{equation}

In summary, the entire MCTS procedure comprises five stages, as illustrated in Algorithm~\ref{alg1}. The final retrieval of relevant nodes is expressed as:
\begin{equation}
\scalebox{0.79}{
$R=MCTS\left( G,\,\,q,\,\,k \right) =\left\{ r_1,\,\,r_2,\,\,...,\,\,r_{\mathrm{k}} \right\} ,\,\,r_{\mathrm{i}}\in V$}
\end{equation}

\section{Experiment}

\subsection{Dataset list}
\label{C.1}

\textbf{QuALITY}\footnote{\href{https://github.com/nyu-mll/quality}{https://github.com/nyu-mll/quality}} is a four-way multiple-choice QA challenge comprising text data from diverse sources, evaluated based on accuracy. The dev set of QuALITY includes 230 long-context samples with corresponding questions, answer options, and other details. 

\textbf{MuSiQue}\footnote{\href{https://github.com/stonybrooknlp/musique}{https://github.com/stonybrooknlp/musique}} serves as a multi-hop long-context QA benchmark, with an average of 15.5 thousand tokens across 200 samples. MuSiQue's questions include various content types, such as narratives, expository texts, and factual material. This diversity challenges models to generalize across various genres and structures, strengthening their capacity for comprehensive language understanding and reasoning.

We also include the \textbf{NarrativeQA}\footnote{\href{https://github.com/google-deepmind/narrativeqa}{https://github.com/google-deepmind/narrativeqa}} dataset as a single-hop long-context benchmark, averaging 29.7 thousand tokens across 200 samples. This dataset focuses on narrative comprehension in a long-form context, requiring models to answer questions based on an understanding of entire narratives, such as books or movie scripts, rather than isolated text segments. This differentiates NarrativeQA from typical QA datasets, which generally focus on shorter text segments.

\subsection{Cost Analysis}
\label{C.2}
Table \ref{tab6} compares the average token consumption across methods used in our experiments. Neural retrieval demonstrates the highest token usage, as it requires converting context chunks into embeddings with the text-embedding-v3 model, adding significant computational overhead. BM25 also has high token consumption, greater than that of the qwen-plus-128k method alone, due to the additional step of evaluating and selecting Top-k chunks for answer generation. Our agent-based method, JERR, consumes 1.23 times more tokens than ReadAgent but achieves superior performance.

\begin{table}[htbp]
\caption{comparison of token consumption per question among all methods on MuSiQue LR evaluation, where "Avg. Ctx. \#Tokens" refers to the average token number of the original dataset. The "Avg. Cost \#Tokens" comprise both input tokens and output tokens during exploration}
\begin{center}
\scalebox{0.63}{
\begin{tabular}{lll}
\hline
Method                          & Avg. Ctx \# Tokens & Avg. Cost \#Tokens \\ \hline
JERR                           & 15.5 k             & 98.54 k            \\
JERR (w.o. Graph Construction) & 15.5 k             & 44.33 k            \\
ReadAgent                       & 15.5 k             & 79.98 k            \\
Neural Retrieval                & 15.5 k             & 138.90 k           \\
BM25                            & 15.5 k             & 39.18 k            \\
qwen-plus-128k                  & 15.5 k             & 16.99 k            \\ \hline
\end{tabular}
}
\label{tab6}
\end{center}
\end{table}

Additionally, once JERR has constructed its graphs, it can reuse these structures efficiently without requiring reconstruction for each new query. This capability significantly reduces JERR's token cost, averaging 44.33k tokens per query when operating without additional graph construction. This efficiency marks a substantial improvement over the token consumption required by ReadAgent.

\section{Differences among JERR, GraphReader and GraphRAG}
\label{D}
GraphReader constructs undirected graphs where nodes represent text chunks or atomic facts. While this captures local relationships, undirected edges lack explicit hierarchical or causal dependencies, leading to inefficient exploration (e.g., cycles or backtracking). While JERR introduces a directed acyclic graph (DAG) to model causal and hierarchical relationships explicitly. For example, edges in JERR’s DAG are weighted by relational attributes (e.g., "causes," "belongs-to"), allowing the agent to prioritize paths with stronger semantic relevance. This structure inherently avoids cycles and supports efficient traversal, whereas GraphReader’s undirected graph requires heuristic rules to prevent redundant exploration.

GraphRAG relies on entity extraction and hierarchical community detection (via Leiden algorithm) to group nodes, followed by community-level summarization. While effective for global query-focused summarization, this approach treats communities as isolated modules, limiting cross-community reasoning. Instead of partitioning nodes into communities, we perform synopsis extraction—an LLM-guided summarization process that preserves logical dependencies across chunks. This generates a condensed yet interconnected set of entities and relationships, avoiding information fragmentation. Unlike GraphRAG’s static community summaries, JERR’s synopses retain hierarchical relationships (via DAGs), enabling multi-hop reasoning across distant text segments.

\end{document}